# Did ChatGPT or Copilot use alter the style of internet news headlines? A time series regression analysis

Chris Brogly
Department of Computer Science
Lakehead University
Orillia, Canada
cbrogly@lakeheadu.ca

Connor McElroy
Department of Computer Science
Lakehead University
Orillia, Canada
csmcelro@lakeheadu.ca

*Abstract*— The emergence of advanced Large Language Models (LLMs) such as ChatGPT and Copilot is changing the way text is created and may influence the content that we find on the web. This study investigated whether the release of these two popular LLMs coincided with a change in writing style in headlines and links on worldwide news websites. 175 NLP features were obtained for each text in a dataset of 451 million headlines/links. An interrupted time series analysis was applied for each of the 175 NLP features to evaluate whether there were any statistically significant sustained changes after the release dates of ChatGPT and/or Copilot. There were a total of 44 features that did not appear to have any significant sustained change after the release of ChatGPT/Copilot. A total of 13 features did appear to have a significant sustained change after the release of ChatGPT and/or Copilot, when using GPT-3 and Gopher's introduction dates as control models that showed no sustained change. A total of 91 other features did show significant change with ChatGPT and/or Copilot although significance with earlier control LLM release dates (GPT-1/2/3, Gopher) removed them from consideration. This initial analysis suggests that these language models may have had only a limited impact on the style of individual news headlines/links, with respect to only some NLP measures.

## I. INTRODUCTION

Large Language Models have transformed how content can be created in recent years due to their capacity to generate realistic texts, stories being one example [1]. Given some well-designed prompts, these models can produce texts that might closely resemble what a human may have written given the same topic [1]. The usage and influence of LLMs across industries with respect to text generation is likely to increase in coming years given easy cloud access to LLMs, and as relevant GPU hardware sees continuing performance increases and potentially lower costs. While there is great potential for automation using this technology, there are some issues that arise from using LLMs in different fields as well. For instance, with regards to journalism, being able to automatically produce headlines, stories, and reporting on different events may at first appear convenient but could present a wide range of new problems related to trust that may result in some additional work or even negative effects for readers. Some of these are considered as follows. First, depending on an LLM to report/summarize current events without sufficient human review could be problematic if some of the generated text is not accurate. Second, LLMs can generate realistic text that is difficult to detect with no training [1] which may result in the proliferation of believable although possibly fake or misleading news [2], which can easily be labelled with a human author even though they are mostly or fully AI generated. Third, LLMs are not always trained using very recent data, which may limit capabilities with regards to text generation for any events that occur outside of the training data date/time range. These are just some considerations with respect to automated text generation using LLMs; there likely are several more that can potentially impact the quality and trustworthiness of news content.

As a result, given some of these potential problems with AI generated texts being incorporated into existing news sources, we were interested in studying whether the releases of major LLMs were associated with any significant changes in various measures of text style on news websites. If there were any significant changes sustained in text style based on a variety of NLP measures after the time of their release, that may indicate that the release of those LLMs coincided with a significant amount of AI-generated text content being published on news websites.

In this work, news headlines and links from a group of worldwide news websites are used with interrupted time series for an initial analysis. With respect to LLMs of interest, ChatGPT and Microsoft Copilot were the main models that were selected here, as ChatGPT is likely the most popular LLM at the time of writing, while Copilot is now directly accessible on every Windows 10/11 desktop installation and is accessible with a browser. The release dates of earlier/less popular models like GPT-3 and Gopher were used as controls in our interrupted time series analysis because GPT-3 was not as widely used as ChatGPT or Copilot and Gopher was not even publicly accessible. Universal Part-of-speech (POS) tags, Penn treebank POS tags, dependency label tags, and Named Entity Recognition (NER) tags were analyzed on each text from the chosen dataset of about 451 million headlines/links resulting in a total of 175 linguistic features. Each of these were analyzed to determine if any sustained significant changes occurred after the release of ChatGPT and/or Copilot.

## II. METHODS

### A. Headline/link dataset information

The dataset used is about 451 million texts consisting primarily of links and headings (full information on tags used can be found in Table 1). This was built from the Common Crawl news dataset over 2016/09/02 to 2023/06/28; every Tuesday and Friday of crawled news pages is included within this range. Common Crawl HTML pages were parsed for hyperlink captions and headings. These captions/headings were then saved to an SQLite database, after which they were passed into another program to obtain 175 NLP features which were saved to a different column in the database. This dataset was used in a previous study on clickbait by our group [3], although all the measurements that are examined here on the texts are new NLP features. The dataset is saved in a standard SQLite database file, with a total size of 489GB consisting of 451,033,388 rows by 195 NLP feature columns. For this analysis, all 195 column values are averaged per day as otherwise this would have taken up a very large amount of RAM and processing time when using standard R packages to fit regression models. Averaged data was used to accommodate for resource limitations; producing this data itself required a system with at least 128GB of RAM and days of processing time.

### B. Feature generation

The 175 NLP features fall into 4 main categories. All of these are generated using the spaCy NLP library using the en_core_web_sm pipeline on a given text from the dataset. These categories include: 1) Universal POS tags, which are commonly found in different NLP libraries. Some examples of these include: "ADJ" (adjective), "VERB", "DET" (determiner). 2) Penn treebank POS tags, which include a larger variety of POS tags, such as "JJS" (adjective, superlative). 3) universal dependency tags, which in this case describe English language grammar. These include tags such as "advmod" (adverbial modifier) or "csubjpass" (clausal subject, passive). 4) Named Entity Recognition (NER) tags which can identify objects in text such as PERSON, WORK_OF_ART, MONEY, etc. All available features for each of the 4 categories are used and recorded per text caption in the database. These were initially saved in CSV format to a single column adjacent to the text caption. They were later moved to separate columns for use in the interrupted time series analysis as described below.

### C. Interrupted time series analysis

An interrupted time series analysis was conducted primarily to determine if any of the 175 NLP features had a statistically significant sustained change following the release of ChatGPT or Copilot. Standard Ordinary Least Squares (OLS) lm() regression models were fit with relevant terms, resulting in a total of 350 time series regression models. The models examined in this work have the same variables as the regression model used in our previous clickbait study [3]; however here we examine 350 of them, or 175 with two event dates (one for Copilot and one for ChatGPT), compared to only 1 regression model with five different events in that previous work. Unlike our previous work, we intentionally chose not to fit Generalized Least Squares (GLS) models (to compensate for autocorrelation) as there are hundreds of models here compared to 1 in the previous work. This is discussed further under limitations.

For this type of interrupted time series [4], there are three variables included with the regression models: T, D, and P. The T (Time Trend) variable represents whether there was a sustained change occurring in advance of the LLM release. Variable D (Event Impact) is a value that captures any change immediately at the LLM release. Variable P (Post-Event Trend) is the estimation of the trend after the event. If P is statistically significant, then that indicates a sustained change each day following the event represented by the P coefficient. This interrupted time series approach is often used for policy changes [4] although we re-purpose it slightly here to analyze the impact of LLM release events.

Although in this paper we focus on ChatGPT and Copilot, initially we wanted to see if any sustained changes occurred with a number of major LLM release dates. The following LLMs were selected at first: GPT-1 (OpenAI), GPT-2 (OpenAI), GPT-3 (OpenAI), Gopher (DeepMind), ChatGPT (OpenAI), Microsoft Copilot (Microsoft), and GPT-4 Turbo (OpenAI). We hypothesized that most P terms related to the earlier LLMs would be non-significant while the later ones might be significant. However, several NLP features showed a significant sustained change after the release of many early models like GPT-1, GPT-2, GPT-3, and even Gopher, which to the best of our knowledge never had a publicly accessible web interface (this result is discussed and considered further under Limitations). It is unlikely that the earlier GPTs influenced writing as they were not nearly as popular or as capable as ChatGPT or Copilot. As a result, we decided to use both GPT-3 (released June 2020) and the unreleased Gopher (announced Dec. 2021) as control models. The idea with the chosen control models is that if a P term came back non-significant for a control, we would have more confidence seeing a significant P term with ChatGPT or Copilot.

## III. RESULTS

First, basic dataset information is provided in Table 1. The results are then split into five tables. For Tables 2-6, the results are shown for NLP features that showed no significant sustained change for ChatGPT and/or Copilot. For Table 7, the results are shown for NLP features that showed no significant sustained change with control models but did show a significant sustained change for ChatGPT and/or Copilot. Shaded cells represent significant p-values ($< 0.05$). Definitions of tags describing table results are used from the spaCy GitHub [5]. Aside from the features displayed in these tables, there were 91 other features which did show significant sustained change with ChatGPT and/or Copilot, but they also had significant sustained changes after the release of the control LLM models

(GPT-1/2/3, Gopher). As a result, these features were removed from consideration and are not reported on here due to our low confidence in them. In comparison, we believe it is more likely that the unchanged NLP features are more accurate given the size of the analyzed dataset.

Table 2 shows all Universal POS tags with no significant sustained change after the release of ChatGPT and/or Copilot. Determiners (DET), pronouns (PRON), subordinating conjunctions (SCONJ), symbols (SYM), and blank space (SPACE) showed no significant sustained change after the release of ChatGPT or Copilot. However, there were some significant changes leading up to their release (T term) and some of them had significant immediate changes (D term).

Table 3 shows all Penn treebank tags with no significant sustained change after the release of ChatGPT and/or Copilot. There are more treebank tags than Universal POS tags. Ones that are not obvious or described previously include: RRB (right round bracket), HYPH (punctuation mark, hyphen), JJS (adjective, superlative), LS (list item marker), PRP (pronoun, personal), PRP$ (pronoun, possessive), RBR (adverb, comparative), TO (infinitival "to"), WP (wh-pronoun, personal), WP$ (wh-pronoun, possessive), and SP (blank space). Again, there are some significant changes leading up to their release (T term) and some immediate changes (D term).

TABLE 1: DATASET PROPERTIES

| Dataset Property | Value |
|---|---|
| Total number of unique news websites analyzed | 26212 |
| HTML tags processed | "a", "span", "h1", "h2", "h3", "h4", "h5", "yt-formatted-string" |
| Hyperlink captions/headings analyzed (sample size/N) | 451,033,388 |
| Minimum word requirement for processed text | 3 |
| Number of days used per feature for time series | 708 |
| Number of part of speech features | 20 |
| Number of Penn treebank features | 57 |
| Number of dependency label features | 72 |
| Number of NER features | 26 |
| Total number of features per text | 175 |

TABLE 2: UNIVERSAL PART OF SPEECH (POS) TAGS WITH LIMITED OR NO SIGNIFICANT SUSTAINED CHANGE (P)

| | ChatGPT (OpenAI) | | | | | | Microsoft Copilot | | | | | |
| Feature | T | Sig? | D | Sig? | P | Sig? | T | Sig? | D | Sig? | P | Sig? |
|---|---|---|---|---|---|---|---|---|---|---|---|---|
| DET | 0.000034 | < 0.001 | -0.024952 | < 0.001 | 0.000034 | 0.0943 | 0.000012 | < 0.001 | -0.017534 | 0.0130 | 0.000029 | 0.2478 |
| PRON | -0.000020 | 0.2944 | -0.000078 | 0.9795 | -0.000020 | 0.0277 | 0.000001 | 0.2912 | -0.001319 | 0.6780 | -0.000021 | 0.0655 |
| SCONJ | 0.000003 | < 0.001 | -0.006496 | < 0.001 | 0.000003 | 0.6082 | 0.000007 | < 0.001 | -0.007384 | < 0.001 | 0.000008 | 0.1999 |
| SYM | -0.000015 | < 0.001 | -0.022018 | < 0.001 | -0.000015 | 0.0386 | 0.000017 | < 0.001 | -0.020208 | < 0.001 | -0.000014 | 0.1173 |
| SPACE | 0.000005 | < 0.001 | -0.002793 | 0.4183 | 0.000005 | 0.6358 | -0.000014 | < 0.001 | 0.000208 | 0.9543 | -0.000002 | 0.8793 |

TABLE 3: PENN TREEBANK PART OF SPEECH TAGS WITH LIMITED OR NO SIGNIFICANT SUSTAINED CHANGE (P)

| | ChatGPT (OpenAI) | | | | | | Microsoft Copilot | | | | | |
| Feature | T | Sig? | D | Sig? | P | Sig? | T | Sig? | D | Sig? | P | Sig? |
|---|---|---|---|---|---|---|---|---|---|---|---|---|
| Period | 0.000006 | 0.0911 | -0.005655 | 0.0219 | 0.000006 | 0.4073 | 0.000001 | 0.1718 | -0.005307 | 0.0406 | 0.000008 | 0.3647 |
| RRB | -0.000033 | < 0.001 | -0.008999 | < 0.001 | -0.000033 | < 0.001 | 0.000025 | < 0.001 | -0.016385 | < 0.001 | -0.000015 | 0.1288 |
| Open quote | -0.000003 | < 0.001 | -0.000595 | 0.5196 | -0.000003 | 0.3495 | -0.000003 | < 0.001 | -0.001454 | 0.1336 | 0.000000 | 0.9092 |
| Close quote | -0.000001 | < 0.001 | -0.000298 | 0.7098 | -0.000001 | 0.5639 | -0.000003 | < 0.001 | -0.000959 | 0.2540 | 0.000000 | 0.9002 |
| DT | 0.000034 | < 0.001 | -0.024858 | < 0.001 | 0.000034 | 0.0998 | 0.000011 | < 0.001 | -0.017342 | 0.0163 | 0.000029 | 0.2598 |
| HYPH | 0.000023 | < 0.001 | 0.004102 | 0.3444 | 0.000023 | 0.0882 | -0.000011 | < 0.001 | 0.008783 | 0.0538 | 0.000012 | 0.4650 |
| JJS | 0.000002 | 0.3081 | 0.001333 | 0.0694 | 0.000002 | 0.3648 | 0.000000 | 0.0447 | -0.000060 | 0.9383 | 0.000006 | 0.0293 |
| LS | 0.000002 | < 0.001 | 0.000390 | 0.1854 | 0.000002 | 0.0603 | 0.000000 | < 0.001 | 0.000459 | 0.1386 | 0.000002 | 0.1304 |
| PRP | -0.000008 | < 0.001 | 0.002279 | 0.0503 | -0.000008 | 0.0197 | -0.000005 | < 0.001 | 0.000692 | 0.5715 | -0.000006 | 0.1479 |
| PRP$ | -0.000006 | < 0.001 | -0.004071 | 0.0123 | -0.000006 | 0.2000 | 0.000006 | < 0.001 | -0.002952 | 0.0847 | -0.000009 | 0.1298 |
| RBR | -0.000001 | 0.2279 | 0.000471 | 0.0590 | -0.000001 | 0.1369 | 0.000000 | 0.6123 | 0.000054 | 0.8353 | 0.000000 | 0.7118 |
| TO | -0.000011 | < 0.001 | -0.006396 | 0.0022 | -0.000011 | 0.0762 | 0.000007 | < 0.001 | -0.003923 | 0.0752 | -0.000018 | 0.0213 |
| WP | -0.000005 | < 0.001 | -0.000263 | 0.5796 | -0.000005 | 0.0013 | 0.000002 | < 0.001 | -0.001944 | < 0.001 | -0.000001 | 0.7151 |
| WP$ | 0.000000 | 0.5319 | -0.000054 | 0.1660 | 0.000000 | 0.0659 | 0.000000 | 0.3125 | -0.000022 | 0.5970 | 0.000000 | 0.2005 |
| SP | 0.000005 | < 0.001 | -0.002792 | 0.4184 | 0.000005 | 0.6361 | -0.000014 | < 0.001 | 0.000206 | 0.9546 | -0.000002 | 0.8793 |

Table 4 shows all dependency label tags with no significant sustained change after the release of ChatGPT and/or Copilot. Dependency labels provide additional information about the structure of texts compared to just POS tags. The dependency labels with no significant sustained changes following the release of ChatGPT and/or Copilot are: advmod (adverbial modifier), agent, attr (attribute), ccomp (clausal complement), compound, csubj (clausal subject), csubjpass (clausal subject, passive), dep (unclassified dependent), det (determiner), meta (meta modifier), npadvmod (noun phrase as adverbial modifier), parataxis, quantmod (modifier of quantifier), relcl (relative clause modifier). There continue to be small significant changes leading up to the release of these models (T term) and often immediate level changes (D term).

Table 5 shows all named entity recognition (NER) tags. These tags are more specific than the previous POS tags with respect to different phrases and words that represent specific entities. The entities with no significant sustained changes after the release of the models included: ROOT, PERSON (people, including fictional), EVENT (named hurricanes, battles, wars, sports events, etc.), LAW (named documents made into laws), LANG (any named language), DATE, TIME, ORDINAL (such as "first").

Table 6 shows all tags that had a significant sustained change (P term) after the models were released, but did not have a significant sustained change after the release of any of the control models. There were often small significant changes before the models were released (T term) and occasionally immediate level changes (D term).

## IV. DISCUSSION

There are a number of interesting findings from this initial statistical analysis of headlines/links on news websites with respect to the release of our selected LLMs. First, at least based on this interrupted time series analysis, a number of POS, dependency label, and NER tags did not appear to have significant sustained changes as a result of ChatGPT and Copilot being launched. Second, there were a limited number of tags that did appear to have significant sustained changes (based on non-significance for control models). Third, our initial results with earlier models produced several significant terms, removing them from consideration. With respect to the first and second findings, given that there are a large number of tags with no significant sustained change and a small number of tags with a significant sustained change, this suggests that the writing style of headlines/links on worldwide news sites was not largely altered after the release of ChatGPT or Copilot, although some of our valid tags with sustained changes like PRP (pronouns) and SYM (symbols) in Table 6 are consistent with other research on text produced by LLMs [6]. With respect to the third finding, where significant sustained increases occurred after the release of earlier, less popular LLMs, there are limitations with this work to discuss. In this analysis of the data, we only used daily averages of each of the 175 NLP features for standard OLS regression models, and did not perform GLS model fits corrected for autocorrelation because of time and memory limitations in both cases. Furthermore, the frequency of significant sustained change for some of the earlier models like GPT-1, 2, and 3 suggests that additional or more advanced statistical analysis would be more precise. As this was web data, we were unable to control for confounding which possibly influenced the results.

TABLE 4: DEPENDENCY LABEL TAGS WITH LIMITED OR NO SIGNIFICANT SUSTAINED CHANGE (P)

| | ChatGPT (OpenAI) | | | | | | Microsoft Copilot | | | | | |
| --- | --- | --- | --- | --- | --- | --- | --- | --- | --- | --- | --- | --- |
| Feature | T | Sig? | D | Sig? | P | Sig? | T | Sig? | D | Sig? | P | Sig? |
| advmod | -0.000003 | < 0.001 | -0.009102 | < 0.001 | -0.000003 | 0.5885 | 0.000009 | < 0.001 | -0.010329 | < 0.001 | 0.000003 | 0.7038 |
| agent | 0.000001 | 0.0273 | -0.001675 | < 0.001 | 0.000001 | 0.4943 | 0.000000 | 0.1612 | -0.001311 | 0.0058 | 0.000001 | 0.6745 |
| attr | 0.000000 | < 0.001 | -0.001533 | 0.0165 | 0.000000 | 0.8683 | 0.000001 | < 0.001 | -0.003142 | < 0.001 | 0.000005 | 0.0434 |
| ccomp | -0.000003 | < 0.001 | -0.007170 | < 0.001 | -0.000003 | 0.3960 | 0.000003 | < 0.001 | -0.005821 | < 0.001 | -0.000004 | 0.2901 |
| compound | 0.000063 | < 0.001 | -0.194844 | < 0.001 | 0.000063 | 0.4405 | 0.000127 | < 0.001 | -0.180567 | < 0.001 | 0.000113 | 0.2687 |
| csubj | 0.000000 | < 0.001 | 0.000337 | 0.0363 | 0.000000 | 0.3280 | 0.000000 | < 0.001 | 0.000813 | < 0.001 | -0.000002 | < 0.001 |
| csubjpass | 0.000000 | 0.1811 | -0.000016 | 0.3897 | 0.000000 | 0.5987 | 0.000000 | 0.1261 | -0.000021 | 0.2906 | 0.000000 | 0.8186 |
| dep | 0.000011 | < 0.001 | -0.007392 | 0.0305 | 0.000011 | 0.3096 | -0.000013 | < 0.001 | -0.003818 | 0.2876 | 0.000005 | 0.7017 |
| det | 0.000032 | < 0.001 | -0.024413 | < 0.001 | 0.000032 | 0.1138 | 0.000012 | < 0.001 | -0.017162 | 0.0175 | 0.000027 | 0.2815 |
| meta | 0.000000 | < 0.001 | 0.000698 | < 0.001 | 0.000000 | 0.8412 | -0.000001 | < 0.001 | 0.000296 | 0.1175 | 0.000001 | 0.1378 |
| npadvmod | 0.000013 | < 0.001 | -0.018202 | < 0.001 | 0.000013 | 0.1712 | 0.000003 | < 0.001 | -0.016646 | < 0.001 | 0.000018 | 0.1153 |
| parataxis | 0.000000 | < 0.001 | 0.000193 | 0.1206 | 0.000000 | 0.6437 | 0.000000 | < 0.001 | 0.000223 | 0.0883 | 0.000000 | 0.9327 |
| quantmod | 0.000002 | < 0.001 | -0.006528 | < 0.001 | 0.000002 | 0.2862 | 0.000005 | < 0.001 | -0.004813 | < 0.001 | 0.000000 | 0.8989 |
| relcl | -0.000004 | < 0.001 | -0.001173 | 0.0919 | -0.000004 | 0.0437 | 0.000001 | < 0.001 | -0.001288 | 0.0783 | -0.000004 | 0.1019 |

TABLE 5: NAMED ENTITY RECOGNITION (NER) TAGS WITH LIMITED OR NO SIGNIFICANT SUSTAINED CHANGE (P)

| | ChatGPT (OpenAI) | | | | | | Microsoft Copilot | | | | | |
| --- | --- | --- | --- | --- | --- | --- | --- | --- | --- | --- | --- | --- |
| Feature | T | Sig? | D | Sig? | P | Sig? | T | Sig? | D | Sig? | P | Sig? |
| ROOT | 0.000009 | < 0.001 | -0.015581 | < 0.001 | 0.000009 | 0.1517 | 0.000002 | < 0.001 | -0.014691 | < 0.001 | 0.000014 | 0.0658 |
| PERSON | -0.000003 | 0.8208 | -0.003514 | 0.3221 | -0.000003 | 0.7559 | 0.000001 | 0.6113 | -0.009614 | 0.0098 | 0.000015 | 0.2498 |
| EVENT | 0.000001 | 0.0469 | -0.000467 | 0.4520 | 0.000001 | 0.5270 | 0.000000 | 0.0325 | -0.000872 | 0.1802 | 0.000003 | 0.2161 |
| LAW | 0.000000 | < 0.001 | -0.000888 | < 0.001 | 0.000000 | 0.6597 | 0.000001 | < 0.001 | -0.000905 | < 0.001 | 0.000001 | 0.3647 |
| LANG | 0.000000 | 0.1905 | -0.000212 | 0.0075 | 0.000000 | 0.5872 | 0.000000 | 0.0446 | -0.000169 | 0.0434 | 0.000000 | 0.6899 |
| DATE | 0.000013 | < 0.001 | -0.029947 | < 0.001 | 0.000013 | 0.3220 | 0.000004 | < 0.001 | -0.031468 | < 0.001 | 0.000032 | 0.0453 |
| TIME | 0.000007 | < 0.001 | 0.005942 | < 0.001 | 0.000007 | 0.1739 | -0.000006 | < 0.001 | 0.006406 | < 0.001 | 0.000005 | 0.4457 |
| ORDINAL | -0.000002 | < 0.001 | -0.002040 | < 0.001 | -0.000002 | 0.1967 | 0.000002 | < 0.001 | -0.001082 | 0.0338 | -0.000004 | 0.0204 |

TABLE 6: ALL TAGS (POS/TREEBANK/DEP. LABEL/NER) WITH SIGNIFICANT SUSTAINED CHANGE (P)

| Feat. | GPT3/ Goph. | ChatGPT (OpenAI) | | | | | | Microsoft Copilot (Microsoft) | | | | | |
|---|---|---|---|---|---|---|---|---|---|---|---|---|---|
| | | T | Sig? | D | Sig? | P | Sig? | T | Sig? | D | Sig? | P | Sig? |
| EX | No/No | 0.000000 | 0.1338 | 0.000263 | < 0.001 | -0.000001 | 0.0022 | 0.000000 | 0.3492 | 0.00021 | 0.0087 | -0.000001 | 0.0051 |
| JJS | No/Yes | 0.000000 | 0.3081 | 0.001333 | 0.0694 | 0.000002 | 0.3648 | 0.000000 | 0.0447 | -0.00006 | 0.9383 | 0.000006 | 0.0293 |
| PRP | No/No | -0.000006 | < 0.001 | 0.002279 | 0.0503 | -0.000008 | 0.0197 | -0.000005 | < 0.001 | 0.00069 | 0.5715 | -0.000006 | 0.1479 |
| WRB | No/No | 0.000004 | < 0.001 | -0.002026 | 0.2014 | 0.000011 | 0.0313 | 0.000004 | < 0.001 | -0.00354 | 0.0327 | 0.000018 | 0.0033 |
| SYM | No/No | 0.000001 | < 0.001 | -0.000972 | 0.3842 | 0.000011 | 0.0020 | 0.000001 | < 0.001 | -0.00060 | 0.6085 | 0.000012 | 0.0047 |
| ADD | Yes/No | 0.000001 | < 0.001 | 0.000820 | < 0.001 | -0.000004 | < 0.001 | 0.000001 | < 0.001 | 0.00113 | < 0.001 | -0.000005 | < 0.001 |
| NFP | Yes/No | -0.000001 | < 0.001 | -0.001289 | < 0.001 | 0.000004 | < 0.001 | -0.000001 | < 0.001 | -0.00113 | 0.0014 | 0.000005 | < 0.001 |
| csubj | No/No | 0.000001 | < 0.001 | 0.000337 | 0.0363 | 0.000000 | 0.3280 | 0.000000 | < 0.001 | 0.00081 | < 0.001 | -0.000002 | < 0.001 |
| expl | No/No | 0.000000 | 0.1718 | 0.000229 | 0.0026 | -0.000001 | 0.0071 | 0.000000 | 0.3177 | 0.00021 | 0.0066 | -0.000001 | 0.0049 |
| Nsubj. pass | No/Yes | -0.000001 | < 0.001 | 0.000402 | 0.4143 | -0.000007 | < 0.001 | -0.000001 | < 0.001 | 0.00036 | 0.4863 | -0.000008 | < 0.001 |
| Num mod | No/Yes | -0.000008 | < 0.001 | -0.018753 | 0.0076 | 0.000140 | < 0.001 | -0.000008 | < 0.001 | -0.01064 | 0.1754 | 0.000150 | < 0.001 |
| ART | No/Yes | 0.000000 | 0.4301 | -0.000402 | 0.3026 | 0.000006 | < 0.001 | 0.000000 | 0.5215 | -0.00031 | 0.4353 | 0.000007 | < 0.001 |
| CARD | No/Yes | 0.000008 | < 0.001 | 0.000354 | 0.8501 | 0.000034 | < 0.001 | 0.000009 | < 0.001 | 0.00005 | 0.9772 | 0.000041 | < 0.001 |

## V. CONCLUSION

This work provides an initial attempt at a large scale interrupted time series analysis regarding the influence of popular LLMs on certain smaller texts (headlines/links) found on worldwide news websites. We argue the use of control models and the size of the dataset (489GB/350 features) provide some confidence in the results. Future work may include more advanced models given our stated limitations and/or an analysis of paragraph text rather than headline/link texts (currently underway by our lab).

ACKNOWLEDGMENT

This work was supported by a grant from Lakehead University, Faculty of Science and Environmental Studies.